\documentclass[lettersize,journal]{IEEEtran}
\usepackage{amsmath,amsfonts}
\usepackage{algorithm}
\usepackage{algpseudocode}
\usepackage{array}
\usepackage[caption=false,font=normalsize,labelfont=sf,textfont=sf]{subfig}
\usepackage{textcomp}
\usepackage{stfloats}
\usepackage{url}
\usepackage{verbatim}
\usepackage{hyperref}
\usepackage{graphicx}
\usepackage{cite}
\usepackage{multicol}
\usepackage{multirow}
\usepackage{cleveref}
\usepackage[normalem]{ulem}
\usepackage{booktabs}
\usepackage{color}
\usepackage{tikz}
\usepackage{MnSymbol}
\usepackage{orcidlink}
\usepackage{authblk}
\graphicspath{{../figs}}

\hypersetup{
  colorlinks   = true, 
  urlcolor     = blue, 
  linkcolor    = blue, 
  citecolor   = blue 
}

\newtheorem{thm}{Theorem}

\newtheorem{prop}{Proposition}

\newcommand{\argmin}{\operatornamewithlimits{argmin}}

\newcommand{\bA}{\bold{A}}

\newcommand{\ba}{\bold{a}}
\newcommand{\bb}{\bold{b}}

\newcommand{\bd}{\bold{d}}

\newcommand{\by}{\bold{y}}
\newcommand{\bx}{\bold{x}}

\newcommand{\bY}{\bold{Y}}
\newcommand{\bX}{\bold{X}}

\newcommand{\bq}{\bold{q}}

\newcommand{\bK}{\bold{K}}

\newcommand{\bC}{\bold{C}}

\newcommand{\bQ}{\bold{Q}}

\newcommand{\bI}{\bold{I}}

\newcommand{\bfbeta}{\boldsymbol{\beta}}
\newcommand{\bfalpha}{\boldsymbol{\alpha}}
\newcommand{\bfrho}{\boldsymbol{\rho}}

\newcommand{\cor}{\operatornamewithlimits{Cor}}

\begin{document}

\title{Systematic Bias of Machine Learning Regression Models and Correction}




\author[1,2,3]{Hwiyoung Lee \orcidlink{0000-0002-3855-2316}} 
\author[1,2,3]{Shuo Chen \orcidlink{0000-0002-7990-4947}}
\affil[1]{Division of Biostatistics and Bioinformatics, Department of Epidemiology and Public Health, School of Medicine, University of Maryland}
\affil[2]{Maryland Psychiatric Research Center, Department of Psychiatry, School of Medicine, University of Maryland}
\affil[3]{The University of Maryland Institute for Health Computing (UM-IHC)}


\maketitle

\begin{abstract}
Machine learning models for continuous outcomes often yield systematically biased predictions, particularly for values that largely deviate from the mean. Specifically, predictions for large-valued outcomes tend to be negatively biased (underestimating actual values), while those for small-valued outcomes are positively biased (overestimating actual values).  We refer to this linear central tendency warped bias as the ``systematic bias of machine learning regression''. In this paper, we first demonstrate that this systematic prediction bias persists across various machine learning regression models, and then delve into its theoretical underpinnings. To address this issue,  we propose a general constrained optimization approach designed to correct this bias and develop computationally efficient implementation algorithms.
Simulation results indicate that our correction method effectively eliminates the bias from the predicted outcomes. We apply the proposed approach to the prediction of brain age using neuroimaging data. In comparison to competing machine learning regression models, our method effectively addresses the longstanding issue of ``systematic bias of machine learning regression'' in neuroimaging-based brain age calculation, yielding unbiased predictions of brain age.

\end{abstract}


\section{Introduction}\label{Sec:Intro}
\IEEEPARstart{C}{onstructing} predictive models with continuous outcomes is a fundamental aspect of modern data science. Numerous tools have been developed for this purpose, including statistical methods such as ordinary linear regression, regression shrinkage, and Generalized Additive Models (GAM), as well as machine learning methods such as random forests, XGBoost, and support vector regression, among others \cite{ESL}. A general objective of these methods is to minimize the discrepancy between the predicted continuous outcomes and the true values, particularly in independent testing datasets. Using this heuristic, the predicted outcome is unbiased under the classic linear regression setting. In contrast, the predicted outcome from a machine learning model for continuous outcomes is often systematically biased \cite{Zhang:2012_RFbias, Belitz:2021}. This systematic bias is problematic for the applications of machine learning regression models, leading to inaccurate conclusions and forecasts. 

We first illustrate the systematic bias introduced by machine learning regression through a simulation study, examining several existing models including Kernel Ridge Regression (KRR), LASSO, XGBoost, Random Forest, Neural Network, and Support Vector Regression (SVR). Specifically, we first generate synthetic training and testing data, each comprising $n=1,000$ observations. The predictors $\bX$ were generated from a multivariate normal distribution with dimension $p=200$ (i.e., $\bX \sim N(0,\Sigma)$, where $\Sigma=I_p$. The response $\by$ is linearly related to the predictor $\bX$, i.e., $\by=\bX\bfbeta + \epsilon$. Here, $\bfbeta \in \mathbb{R}^p$ is a vector of regression coefficients, with each coefficient set to 0.1, and $\epsilon$ denotes random noise, which follows the standard normal distribution.

\begin{figure*}[ht!]
    \centering
    \includegraphics[width=0.99\linewidth]{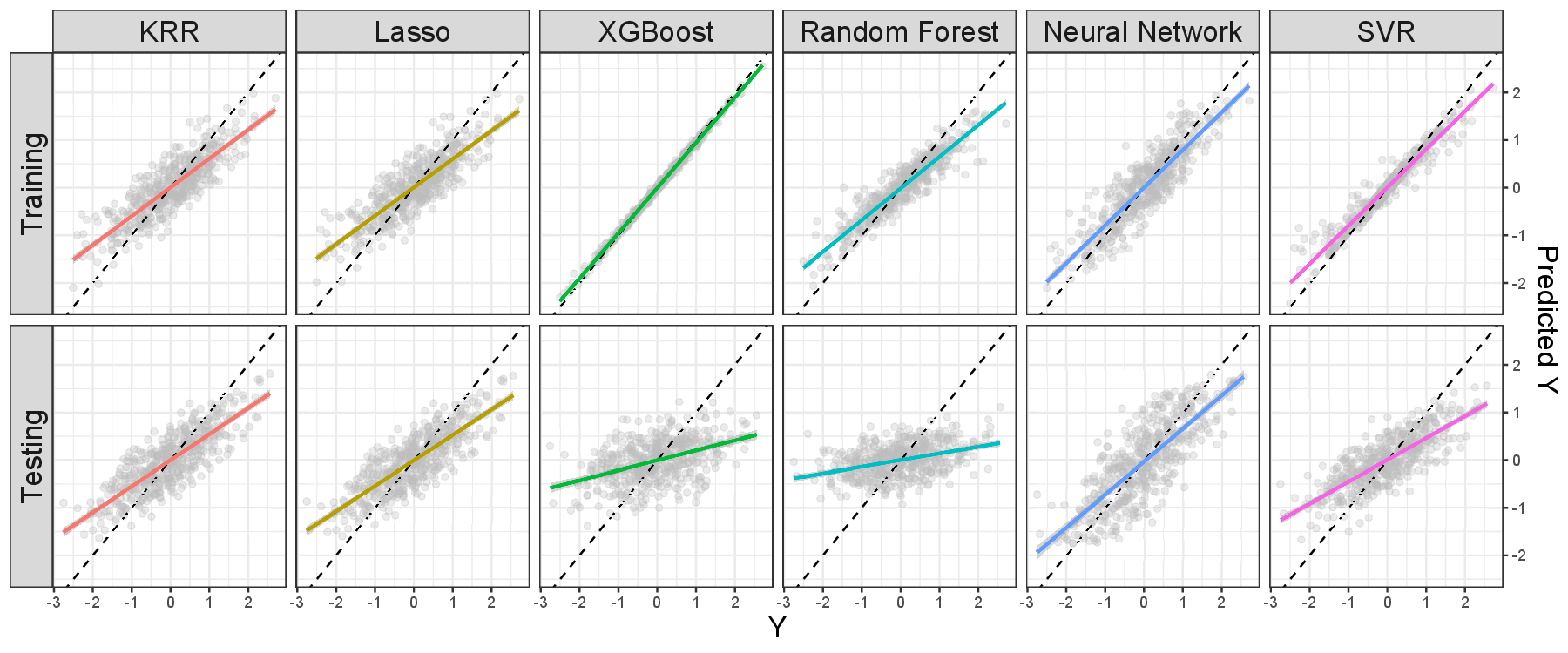}
    \caption{The systematic bias of machine learning regression models. The simulated training and testing set each includes $n=1,000$ observations of $p=200$ features. Six machine learning methods (shown in grey panels), including Kernel Ridge Regression (KRR), LASSO Regression, XGBoost, Random Forest, Neural Network, and Support Vector Regression (SVR), are used. In each scatter plot, each dot represents a pair of predicted outcome $\widehat{\by}_i$ and true outcome ${y}_i$ and the solid line is the regression line of $\widehat{\by}_i$ and ${\by}_i$. The dotted line has a slope of 1 where $\widehat{\by}_i$ exactly equals to ${\by}_i$. The machine learning regression is biased when the solid line deviates from the dotted line. All six machine learning models of this simulation analysis demonstrate a systematic bias.}
    \label{fig:MTM_bias}
\end{figure*}

\Cref{fig:MTM_bias} presents  scatter plots of the predicted outcome values $\widehat{\by}$ vs. true outcomes $\by$ by the aforementioned machine learning regression models based on our simulation analysis. The solid line in each subfigure represents the linear regression fit of the predicted values $\widehat{\by}$ to the true values $\by$. Additionally, a dotted line with a slope of 1 is included, representing the line where $\widehat{\by} = \by$. If the machine learning regression is unbiased, the solid line (i.e., the regression of $\widehat{\by}$ on $\by$) should coincide with the dotted line. As shown in this figure, a systematic regression error is observed across all machine learning regression models. Specifically, there is a tendency to overestimate when the true value of $\by$ is small and to underestimate when the true value of $\by$ is large. Although the sum of residuals $\widehat{\by} - \by$ is approximately zero across all observations, the center-warping tendency presents a ubiquitous systematic bias for machine learning regression models.
This systematic bias is more pronounced in predicted outcomes of the testing dataset.

Since the systematic bias shows a consistent center-warping and linear tendency across all machine learning models, we refer to this phenomenon as the ``systematic bias of machine learning regression'' (SBMR). We denote the linear trend of this systematic bias by $c=\mbox{sin}(\theta)$, where $\theta$ is the angle between the dotted line and the solid line. In the following \Cref{lem1}, we show that the mean squared error of the biased prediction from a machine learning regression model is smaller than the unbiased prediction. This theoretically explains why the aforementioned systematic bias is preferred by all machine learning regression models with the objective function to minimize the mean squared error.  

\begin{prop}\label{lem1}
Consider the outcome $Y$ with a mean of zero and variance of $\sigma^2$, $\ddot{\by}_i$ as the unbiased prediction  of $\by_i$ ($i= 1,\cdots, n$), and the systematically biased machine learning regression prediction $\tilde{y_i}=\ddot{\by}_i-c\ddot{\by}_i$, where $0<c<1$. Then for $c$, and the coefficient of determination $R^2$ of $\ddot{\by}$ satisfying $c<\frac{2-2R^2}{2-R^2}$, we have
\begin{align*}
    \mathbb{E}\left[\sum_{i=1}^n\left(\by_i-\ddot{\by}_i\right)^2\right] \geq \mathbb{E}\left[\sum_{i=1}^n\left(\by_i-\tilde{\by}_i\right)^2\right].
\end{align*}
\end{prop}

This can be straightforwardly checked by using $\mathbb{E}[\sum_{i=1}^n\left(\by_i-\ddot{\by}_i\right)^2] = (n-1)\sigma^2(1-R^2)$, and $\mathbb{E}[\sum_{i=1}^n\left(\by_i-\tilde{\by}_i\right)^2] = (n-1) c^2\sigma^2 + (n-1)(1-c)^2\sigma^2(1-R^2)$.

The above proposition shows that predicted outcomes $\widetilde{\by}$ from machine learning regression models with a systematic bias generally has a smaller than outcomes $\widehat{\by}$ of unbiased  machine learning regression models. Since machine learning regression models are designed to minimize the squared error (i.e., variance), the bias naturally increases due to the `bias-variance' trade-off. However, this issue is problematic when increased bias is systematic (see \Cref{fig:MTM_bias}) because the systematic bias leads to suboptimal decision making and inaccurate forecasts. 

Recently, this issue garnered increased interest in the neuroimaging community, especially in the context of calculate brain age, a biological age characterizing the brain health condition, based on machine learning models with chronological age as the continuous outcome and high-throughput neuroimaging data as predictors. Numerous studies have attempted to tackle the systematic bias issue. Most proposed solutions focus on performing a post-bias correction step. For example, \cite{SMITH:2019} introduced a two-stage method that involves regressing residuals on the response after the initial model fitting. However, as highlighted by \cite{Butler:2021}, the post-bias correction carries associated risks, particularly in making the corrected brain age predictions more dependent on chronological age rather than neuroimaging predictors. For example, if there is no relationship between the chronological age and the neuroimaging predictors, the correction process result in a brain age solely determined by chronological age. Furthermore, as noted by \cite{Treder:2021, Wang:2023} this post-bias correction is problematic within a predictive modeling framework because it essentially involves adjusting the response variable $\by$ while model's predictions to remain biased. This violates the foundational principles of predictive modeling, where the response of the testing set is assumed to be unseen. 
Therefore, there is unmet need of systematic-bias correction for machine learning regression models in both neuroimaging community and machine learning community. To fill this gap and tackle the SBMR issue, we propose a novel approach that resolves the problem by proposing new objective functions for unbiased machine learning regressions, without requiring any subsequent correction steps. This is achieved by imposing constraints designed to prevent the fitted values from being forced toward the mean.

The rest of the paper is organized as follows. In \Cref{Sec:method}, we propose constraints that prevent systematic bias in predicted values and introduce linear and nonlinear regression models incorporating these constraints. In \Cref{Sec:Simulation}, we evaluate the proposed method by comparing it with other machine learning methods. In \Cref{Sec:Real}, we apply the proposed methods to estimate brain age using two neuroimaging datasets.

\section{Method}\label{Sec:method}

In this section, we present a general approach for bias correction in machine learning regression models. Specifically, we propose new constraints to the objective function, which ensure unbiased predictions.

We denote the overall mean of the response variable as $\overline{\by} = \frac{1}{n}\sum_{i=1}^n \by_i$ for the training dataset.  The response variable can then be categorized into two groups based on the overall mean: (i) those less than the overall training mean, i.e., $\mathbb{I}_{<} = \{i \in n \vert \by_i < \overline{\by} \}$, and (ii) those greater than the overall training mean, i.e., $\mathbb{I}_{>} = \{i \in n \vert \by_i > \overline{\by} \}$. The corresponding training means for each group are $\overline{\by}_{<} = \frac{1}{\vert \mathbb{I}_{<} \vert} \sum_{i \in \mathbb{I}_{<}} \by_i$, and $\overline{\by}_{>} = \frac{1}{\vert \mathbb{I}_{>} \vert} \sum_{i \in \mathbb{I}_{>}} \by_i$.

As illustrated in \Cref{fig:MTM_bias}, there is a trend of overestimation to the left of the overall mean and underestimation to the right. Therefore, we propose a new constrained objective function that automatically corrects the systematic prediction bias. Specifically, we focus separately on the two sets $\mathbb{I}{<}$ and $\mathbb{I}{>}$, with our proposed constraints designed to ensure that the sum of residuals $\sum_i(\widehat{\by}_i - \by_i)$ within each set equals zero. We further denote the mean value of the predicted outcomes for $\mathbb{I}_{<}$ in the training dataset as $\texttt{Avg}(\widehat{\by}_{<}) = \frac{1}{\vert \mathbb{I}_{<}\vert} \sum_{i\in\mathbb{I}_{<}}\widehat{\by}_i$, and $\texttt{Avg}(\widehat{\by}_{>}) = \frac{1}{\vert \mathbb{I}_{>}\vert} \sum_{i\in\mathbb{I}_{>}}\widehat{\by}_i$ for $\mathbb{I}_{>}$.
Jointly, we have the constraints as follows:
\begin{equation}
\begin{aligned}
    \overline{\by}_{<} - \texttt{Avg}(\widehat{\by}_{<}) = 0 \\
    \overline{\by}_{>} - \texttt{Avg}(\widehat{\by}_{>})  = 0.
\end{aligned}
\label{eq:constraints_general}
\end{equation}

These constraints enforce that the sum of the discrepancies between the actual values $\by_i$ and predicted values of outcomes $\widehat{\by}_i$ are zero for both subsets $\mathbb{I}_{<}$ and $\mathbb{I}_{>}$. 
In the following \textit{\Cref{thm:constraint}}, we show that implementing the above constraints can correct the systematic bias of the predicted outcomes by machine learning regression models.


\begin{thm}\label{thm:constraint}
    For response $\bY$ with any distribution, let $\widetilde{\by}_i$ denotes the prediction of $\by_i$, and suppose it has the form $\tilde{\by}_i=\ddot{\by}_i-c\ddot{\by}_i$, where $\ddot{\by}$ is an unbiased prediction. If two constraints in \eqref{eq:constraints_general} hold, then systematic bias $c=0$ (\textit{see proof in the Appendix}). 
\end{thm}

Thus, these constraints can be applied to correct the overestimation bias associated with lower response values, and the underestimation bias observed to the right of the overall mean. To ensure unbiased predictions, we impose the following constraints on the objective function of machine learning regression models. Next, we implement the optimization of machine learning regression models including LASSO and KRR with the proposed constraints for unbiased prediction.

\subsection{LASSO with unbiased prediction (LASSO-UP)}
We begin by introducing the methods for implementing the proposed constraints on a commonly used regression shrinkage model. The regularized linear regression takes the following form: 
\begin{align}
    \argmin_{\bfbeta\in\mathbb{R}^p} \sum_{i=1}^n \left(\by_i -\bx_i^\top\bfbeta\right)^2 + \lambda \Omega(\bfbeta),
\end{align}
where $\Omega$ is a penalty function. For the demonstration purpose, we adopt the popular model of LASSO regression as an example \cite{Tibshirani:1996}, which utilizes the $\ell_1$ norm penalty for $\Omega$, i.e., $\Vert \bfbeta \Vert_1 = \sum_{j=1}^p \vert \bfbeta_j \vert$. As with the response variable, the predictors can be divided into two distinct sets: $\bX_< = [\bx_i : i \in \mathbb{I}_{<}] \in \mathbb{R}^{\vert\mathbb{I}_{<}\vert \times p}$, and $\bX_> = [\bx_i : i \in \mathbb{I}_{>}] \in \mathbb{R}^{\vert\mathbb{I}_{>}\vert \times p}$.



Under the linear model, where the prediction for the $i$-th subject can be calculated as $\widehat{\by}_i = \bX_i\bfbeta$, the constraints \eqref{eq:constraints_general} can be expressed as linear equality constraints in a matrix form. 

Specifically, we refer to the LASSO optimization with proposed constraints as LASSO with unbiased prediction (LASSO-UP). The updated objective function and constraints of LASSO-up are:
\begin{alignat*}{2}
    &\argmin_{\bfbeta\in\mathbb{R}^p} \quad && \frac{1}{2}\Vert \by-\bX\bfbeta\Vert^2 + \lambda \Vert \bfbeta \Vert_1\\
    &\text{subject to} \quad &&
    \underbrace{
    \begin{pmatrix}
        \bf{1}_{\vert \mathbb{I}_{<} \vert}^\top \bX_{<} \\
        \bf{1}_{\vert \mathbb{I}_{>} \vert}^\top \bX_{>}
      \end{pmatrix}}_{\bC_\bX \in \mathbb{R}^{2\times p}}
    \bfbeta = 
    \underbrace{
        \begin{pmatrix}
        \vert \mathbb{I}_< \vert \cdot \overline{\by}_< \\
        \vert \mathbb{I}_> \vert \cdot \overline{\by}_>
    \end{pmatrix}}_{\bb_{\by}\in \mathbb{R}^{2}}
\end{alignat*}
The above form is the constrained LASSO problem \cite{Gaines:2018} with an equality-only constraint. In practice, it is preferable to reformulate this problem as a quadratic programming (QP), thereby enabling more straightforward optimization. Specifically, $\bfbeta$ can be decomposed into $\bfbeta=\bfbeta^{+}-\bfbeta^{-}$, where $\bfbeta^+_i=\bfbeta_i I(\bfbeta_i>0)$, and $\bfbeta^{-}_i=\vert\bfbeta_i\vert I(\bfbeta_i<0)$. As this decomposition can remove the absolute operator (i.e., $ \Vert \cdot \Vert_1$) in the original LASSO objective function, We can thus derive the following general QP formulation:
\begin{align*}
   &\argmin_{(\bfbeta^{+},\bfbeta^{-})\in\mathbb{R}^{2p}} &&\frac{1}{2} \begin{pmatrix} \bfbeta^{+}\\ \bfbeta^{-}\end{pmatrix}^\top
   \begin{pmatrix} \bX^\top\bX & -\bX^\top\bX\\ -\bX^\top\bX & \bX^\top\bX \end{pmatrix}
   \begin{pmatrix} \bfbeta^{+}\\ \bfbeta^{-}\end{pmatrix}\\
   & &&+ \left(\lambda \bI_{2p} - \begin{pmatrix} \bX^\top\by \\ -\bX^\top\by \end{pmatrix}\right)^\top \begin{pmatrix} \bfbeta^{+}\\ \bfbeta^{-}\end{pmatrix}\\
   &\text{subject to} && \left(\bC_{\bX} -\bC_{\bX}\right)\begin{pmatrix} \bfbeta^{+}\\ \bfbeta^{-}\end{pmatrix} = \bb_{\by}\\
   & && \bfbeta^{+} \geq \bold{0}_p, \bfbeta^{-1} \geq \bold{0}_p.
\end{align*}

To derive a numerical algorithm for solving LASSO-UP, we first establish some notations. We define $\bA \in \mathbb{R}^{(2+2p)\times 2p}$ by row-wise concatenation of matrices $\bC$ and $-\bI_{2p}$, i.e., $\bA = \begin{bmatrix}\bC\\-\bI_{2p}\end{bmatrix}$, where $\bC=\left(\bC_{\bX}, -\bC_{\bX}\right)\in\mathbb{R}^{2\times 2p}$, and $\bI_{2p}$ is an identity matrix of size $2p$.
We also define $\widetilde{\bfbeta}= ({\bfbeta^+}^\top, {\bfbeta^-}^\top)^\top \in \mathbb{R}^{2p}$, and $\bb = {({\bb_{\by}}^\top, {\bold{0}_{2p}}^\top)}^\top \in \mathbb{R}^{2+2p}$. Then the constraint term is formulated as the following linear equation: $\bA\widetilde{\bfbeta} = \bb$. We also let $\bQ$ be the matrix in the quadratic term, and $\bq$ be the matrix in the linear term of the objective function. Using these definitions, we express the Lagrangian for our problem as follows:
\begin{align*}   
\mathcal{L}(\widetilde{\bfbeta},\bfrho)=\frac{1}{2}\widetilde{\bfbeta}^\top \bQ \widetilde{\bfbeta} + \bq^\top \widetilde{\bfbeta} + \sum_{i=1}^2 \bfrho_i\underbrace{(\ba_i^\top\widetilde{\bfbeta}-\bb_i)}_{(i)} +  \sum_{j=3}^{2+2p}\bfrho_j \underbrace{(\ba_j^\top\widetilde{\bfbeta}-\bb_j)}_{(ii)}
\end{align*}
, where $\bfrho=\mathbb{R}^{2+2p}$ is a vector of Lagrangian multipliers, and $\ba_i\in\mathbb{R}^{2p}$ denotes the $i$-th row of $\bA$, and $\bb_i$ denotes the $i$-th element of $\bb$. In the above Lagrangian function, $(i)$, and $(ii)$ correspond to the equality and nonnegativity constraints, respectively. We denote the set of equality constraints by $\mathcal{E}$ and the set of inequality constraints by $\mathcal{I}$, respectively.

Now this can be optimized by utilizing QP techniques, such as the active set method \cite{Nocedal:2006}, which iteratively optimizes the objective function by selectively updating and solving under a subset of constraints. In the active set framework, this specific subset of constraints is referred to as the working set, denoted as $\mathcal{W}_k$ at the $k$-th iteration. The working set, $\mathcal{W}_k$, includes all equality constraints, $\mathcal{E}$, and some active inequality constraints, represented as $\mathcal{W}_k = \mathcal{E} \cup \{i : \mathbf{a}_i^\top \widetilde{\boldsymbol{\beta}}_k = b_i, i \in \mathcal{I}\}$.
The general form of the active-set method for solving quadratic programming (QP) problems is summarized in \Cref{algo:const_Lasso}. For more detailed information on the active-set method, see \cite{Nocedal:2006}.


\begin{algorithm}[ht!]
\caption{Algorithm for constrained LASSO-UP}\label{algo:const_Lasso}
        \begin{algorithmic}[1]
                \State Start with the initial feasible point $\widetilde{\bfbeta}_0$
                \State Find the initial working set $W_0$
                \For{$k=0,1,2\cdots$}
                \State Compute the gradient of the objective function at the current point:
                \begin{align*}
                    \nabla f (\widetilde{\bfbeta}_k) = \bQ\widetilde{\bfbeta}_k + \bq
                \end{align*}
                \State Compute the search direction $\bd_k$ by solving the linear system
                    \begin{align*}
                    \begin{bmatrix}
                        \bQ & \bA_k^\top\\
                        \bA_k & 0
                    \end{bmatrix}
                    \begin{bmatrix}
                        \bd_k\\
                        \bfrho
                    \end{bmatrix}=
                    \begin{bmatrix}
                        -\nabla f (\widetilde{\bfbeta}_k)\\
                        0
                    \end{bmatrix}                    
                    \end{align*},
                    where $\bA_k$ having the rows of $\ba_i, i \in \mathcal{W}_k$
                    \If{$\bd_k=0$}
                        \State Check optimality:
                            \State Calculate Lagrangian multipliers $\bfrho_i$
                            \begin{align*}
                                \sum_{i\in\mathcal{W}_k}\ba_i\bfrho_i = \nabla f (\widetilde{\bfbeta}_k)
                            \end{align*}
                        \If{$\bfrho_i \geq 0$ for $\forall i \in \mathcal{W}_k \cap \mathcal{I}$}
                           \State \textbf{stop} with solution $\widetilde{\bfbeta}_\ast=\widetilde{\bfbeta}_k$
                        \Else
                            \State $j \leftarrow \argmin_{i \in \mathcal{W}_k \cap \mathcal{I}} \bfrho_j$
                            \State $\mathcal{W}_{k+1} \leftarrow \mathcal{W}_k \backslash \{j\}$
                            \State $\widetilde{\bfbeta}_{k+1} \leftarrow \widetilde{\bfbeta}_{k}$
                        \EndIf
                    \Else \ {$\bd_k\neq 0$}
                        \State Compute the step size $s_k$
                            \begin{align*}
                                s_k = \min_{i\notin \mathcal{W}_{k}, \ba_i^\top\widetilde{\bfbeta}_k>0}\left(1,\frac{\bb_i-\ba_i^\top \widetilde{\bfbeta}_k}{\ba_i^\top\bd_k}\right)
                            \end{align*}
                        \State Update $\widetilde{\bfbeta}_{k+1} \leftarrow \widetilde{\bfbeta}_{k} + s_k \bd_k$
                        \If{$s_k < 1$}
                        \State Find one blocking constraint $i\notin \mathcal{W}_{k}$
                        \State $\mathcal{W}_{k+1} \leftarrow \mathcal{W}_k \cup \{i\}$
                        \Else 
                        \State $\mathcal{W}_{k+1} \leftarrow \mathcal{W}_k$
                        \EndIf
                    \EndIf
            \EndFor
        \end{algorithmic}
\end{algorithm}



\subsection{KRR with unbiased prediction (KRR-UP)}
We further apply the proposed constraints to nonlinear models to achieve predictions without systematic bias. We denote  a general nonparametric regression by $\by_i = f(\bx_i) + \epsilon_i$, where the goal is to estimate a function $\widehat{f}$ from a function space $\mathcal{F}: \mathcal{X} \rightarrow \mathbb{R}$. One popular choice for the function space $\mathcal{F}$ is the Reproducing Kernel Hilbert Space (RKHS), denoted by $\mathcal{H}$. Given that $\mathcal{H}$ is infinite-dimensional, regularization is essential for estimation. The optimization problem known as kernel ridge regression (KRR) \cite{ScSm:2002} can be formulated as follows:
\begin{align*}
   \widehat{f} = \argmin_{f\in\mathcal{H}} \sum_{i=1}^n \left(\by_i -f(\bx_i)\right)^2 + \lambda \Vert f\Vert_{\mathcal{H}}^2,
\end{align*}
where the first term is goodness-of-fit to data, and the second term is the regularization term which controls the trade-off between the fit of the model to the data and the smoothness of the function $f$.
By the representer theorem \cite{Wahba:1990}, the function $f$ can be expressed as a linear combination of kernel functions, i.e.,  $f(\bx) = \sum_{i=1}^n \alpha_i \kappa (\bx,\bx_i)$. The function $\kappa: \mathcal{X} \times \mathcal{X} \rightarrow \mathbb{R}$ is called a kernel function if it is symmetric and positive definite. The most popular choice of the kernel is the Gaussian Radial Basis Function (RBF) Kernel, i.e., $\kappa(\bx_i,\bx_j) = \exp(-\Vert \bx_i-\bx_j\Vert^2/\sigma^2)$.

By imposing the two equality constraints in \eqref{eq:constraints_general}, the KRR objective function can be expressed as follows:
\begin{alignat}{2}
    \widehat{f} = &\argmin_{f\in \mathcal{H}} &&\sum_{i=1}^n \left(\by_i -f(\bx_i)\right)^2 + \lambda \Vert f\Vert_{\mathcal{H}}^2\notag\\
        &\text{subject to} \quad && \sum_{i\in \mathbb{I}_{<}}  \overline{\by}_< -f(\bx_i)  = 0 \label{eq:constrained_KRR_original}\\ 
        &  && \sum_{i\in \mathbb{I}_{>}}  \overline{\by}_{>} -f(\bx_i)  = 0\notag
\end{alignat}

One of the primary advantages of the kernel method is its ability to apply nonlinear techniques within a linear algorithmic framework by utilizing the kernel trick \cite{Hofmann:2008}. Consequently, unlike other nonlinear regression methods, KRR benefits from the incorporation of the mean equality constraints, which can be treated as linear constraints with respect to the optimization variables. Specifically, by
substituting the representation of $f(\cdot)$ into \eqref{eq:constrained_KRR_original}, it can express in a dual form:
\begin{alignat*}{2}
    \widehat{\bfalpha}=&\frac{1}{2}\argmin_{\alpha\in\mathbb{R}^n} && \Vert \by-\bK\bfalpha \Vert_2^2 + \lambda \bfalpha^\top \bK \bfalpha \\
    &\text{subject to} \quad &&
    \underbrace{
    \begin{pmatrix}
        \bf{1}_{\vert\mathbb{I}_<\vert}^\top \bK_{<} \\
        \bf{1}_{\vert\mathbb{I}_>\vert}^\top \bK_{>}
    \end{pmatrix}}_{\bC_{\bK}}
    \bfalpha= 
        \begin{pmatrix}
        \vert \mathbb{I}_{<} \vert \cdot \overline{\by}_< \\
        \vert \mathbb{I}_{>} \vert \cdot \overline{\by}_>
    \end{pmatrix},
\end{alignat*}
where $\bK\in\mathbb{R}^{n\times n}$ denotes the kernel gram matrix of the complete data with the $(i,j)$ element is $\kappa(\bx_i,\bx_j)$, and  $\bK_{<}\in\mathbb{R}^{\vert \mathbb{I}_{<} \vert\times n}$, $\bK_{>}\in\mathbb{R}^{\vert \mathbb{I}_{>} \vert\times n}$ denote the kernel matrices corresponding to datasets $\mathbb{I}_{<}$, $\mathbb{I}_{>}$, respectively. Thus, for example the $i$-the row of $\bK_{<}$ is $\bK_{{<}(i,\cdot)}=\left(\kappa(\bx_{{<}_i},\bx_1),\cdots,\kappa(\bx_{{<}_i},\bx_n)\right)$, where ${\mathbb{I}_{<}}_i$ denotes the $i$-th subject in the set $\mathbb{I}_{<}$. 
Optimizing the main objective function is equivalent to minimizing:
\begin{align*}
\widehat{\bfalpha}=&\argmin_{\alpha\in\mathbb{R}^n} && \frac{1}{2}\bfalpha^\top(\bK+\lambda\bI)\bfalpha-\by^\top\bfalpha\\
&\text{subject to} && \bC_{\bK}\bfalpha = \bb_{\by} 
\end{align*}

The optimization problem described is an equality-constrained quadratic programming (EQP), which consists of a quadratic objective function and a linear equality constraint, both in the optimization variable, $\bfalpha$. Since the optimization problem is convex, it can be readily solved using the method of Lagrange multipliers. To derive the numerical algorithm, we express the objective function in Lagrangian form as follows:
\begin{align*}
    \mathcal{L}(\bfalpha,{\boldsymbol{\rho}}) = &\frac{1}{2}\bfalpha^\top \left(\bK+\lambda\bf{I}\right)\bfalpha-\by^\top\bfalpha+{\boldsymbol{\rho}}^\top \left(\bC_{\bK}\bfalpha-\bb_{\by}\right),
\end{align*}
where $\bfrho \in \mathbb{R}^2$ is a vector of Lagrangian multipliers.
The first-order necessary condition for the optimalities is:
\begin{align*}
    \nabla_{\bfalpha} \mathcal{L}&= (\bK+\lambda\bf{I})\bfalpha-\by + {\bC_{\bK}}^\top \bfrho=0\\
    \nabla_{\bfrho} \mathcal{L} &=  {\bC_{\bK}}\bfalpha-\bb_{\by}=0.
\end{align*}
Thus, it can be optimized by sequentially updating $\bfalpha$ and $\bfrho$ using gradients. We summarize the numerical algorithm for solving KRR-UP in \Cref{algo:const_KRR}. 

\begin{algorithm}[ht!]
\caption{Algorithm for KRR-UP}\label{algo:const_KRR}
        \begin{algorithmic}[1]
            \State Initialize $\bfalpha_0$ and $\bfrho_0$ with feasible starting 
            \For{$k=0,1,2,\cdots$}
                \State Update $\bfalpha_{k}$:
                \State \hspace{1em} Find the search direction (gradient)
                \begin{align*}
                    \nabla_{\bfalpha} \mathcal{L} (\bfalpha_{k},\bfrho_k) = (\bK+\lambda{\bf{I}}){\bfalpha}_k-\by + {\bC_{\bK}}^\top {\bfrho}_k
                \end{align*}
                \State \hspace{1em} Gradient step
                \begin{align*}
                    \bfalpha_{k+1} = \bfalpha_{k} -s_k \nabla_{\bfalpha} \mathcal{L} (\bfalpha_{k},\bfrho_k) 
                \end{align*}
                \State Update $\bfrho_{k}$:
                \State \hspace{1em} Find the search direction (gradient)
                \begin{align*}
                    \nabla_{\bfrho} \mathcal{L} (\bfalpha_{k+1},\bfrho_k) =  {\bC_{\bK}}\bfalpha_k-\bb_{\by}
                \end{align*}
                \State \hspace{1em} Gradient step 
                \begin{align*}
                    \bfrho_k =  \bfrho_k + t_k  \nabla_{\bfrho} \mathcal{L} (\bfalpha_{k+1},\bfrho_k)
                \end{align*}
            \EndFor\\
                Note: $s_k$ and $t_k$ denote the step size at each iteration $k$, which can be updated by the line search methods \cite{Nocedal:2006}, or can be fixed (i.e., $s_k=s$ and $t_k=t$).
        \end{algorithmic}
\end{algorithm}

\begin{figure}[ht!]
    \centering
    \includegraphics[width=0.99\linewidth]{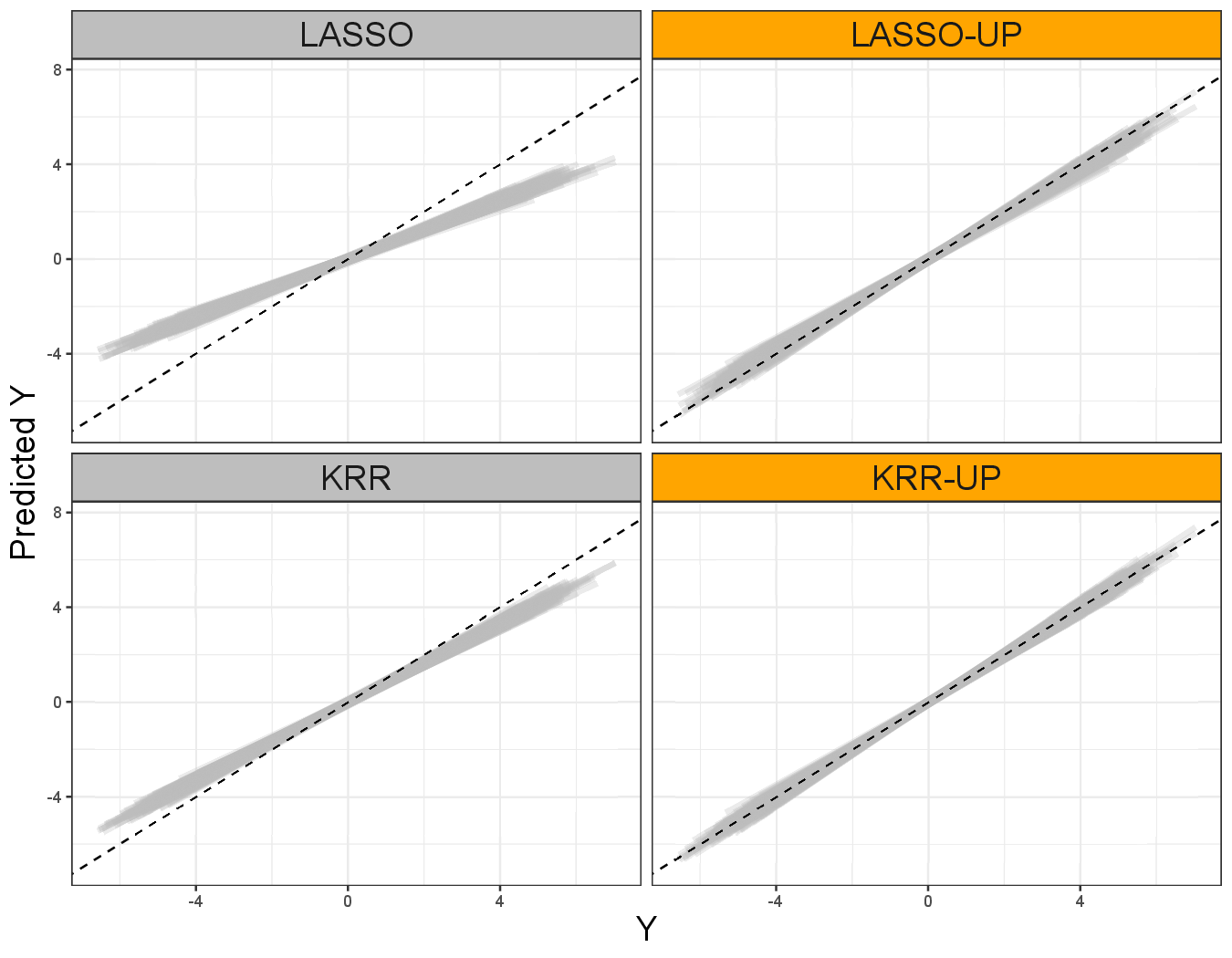}
    \caption{Using the simulation setup described in \Cref{Sec:Intro}, we conducted 500 replications and displayed regression lines between the predicted response value and the true response for each replication; each line is indicated in gray. The black dashed line represents the reference line with a slope of 1 (i.e., $\widehat{\by}=\by$), serving as a benchmark for ideal predictions. The methods with constraints (shown in orange panels) correct prediction bias, whereas the unconstrained versions display a clear systematic bias in predictions.}
    \label{fig:Constraint}
\end{figure}

To illustrate how these constraints correct prediction bias, we compare the LASSO-UP and KRR-UP, with their unconstrained counterparts through simulations. \Cref{fig:Constraint} displays regression line plots of the predicted versus true responses, replicated 500 times, demonstrating the absence of systematic bias when constraints are imposed on the models (LASSO and KRR).

\section{Simulation}\label{Sec:Simulation}
In this section, we benchmark our proposed methods (KRR-UP, and LASSO-UP) against their unconstrained counterparts (KRR, and LASSO \cite{Tibshirani:1996}) as well as other machine learning methods, including XGBoost \cite{Chen:2016}, Random Forest (RF; \cite{Breiman2001}), Neural Network (NN; \cite{Schmidhuber:2015}), and Support Vector Regression (SVR; \cite{SVR:1996,Smola:2004}). The implementations of these methods were carried out using the following \texttt{R} packages: \texttt{xgboost} for XGBoost, \texttt{randomForest} for Random Forest, \texttt{neuralnet} for Neural Networks, and \texttt{e1071} for SVR. We consider both linear and nonlinear relationships between predictors $\bX$ and response $\by$. In the linear case, we first generate $\bX$ from the multivariate normal distribution with dimension $p=10$ and then generate the response $\by = \bX\bfbeta + \epsilon$, where $\epsilon \sim N(0,\sigma_e^2)$. In the nonlinear case, we generate $\by = f(\bX) + \epsilon$, and for the form of $f(\bX)$, we use two of Friedman's test functions \cite{Friedman:1991}, which are commonly used for evaluating machine learning algorithms:
\begin{align*}
        1. f(\bX) = &0.1\exp(4\bX_1) + 4/(1+\exp(-20(\bX_2-0.5))\\
        & +3\bX_3 + 2\bX_4 +\bX_5 + 0\cdot\sum_{p=6}^{10}\bX_p\\
        2. f(\bX) = &10\sin(\pi \bX_1 \bX_2) + 20(\bX_3-0.5)^2 + 10\bX_4 + 5\bX_5
\end{align*}

We fitted the models using only the training set and evaluated its performance on both the training set and the testing set, which was unseen during the training stage. For all settings, we use an equal number of samples for the training and testing sets, with each set containing $n=100$.

We evaluate the performance of methods using the following criteria.
To assess the systematic bias of predictions by these machine learning regression models, we evaluate bias by comparing the regression slope—obtained by regressing predicted values on observed values—to the reference line with a slope of 1 (i.e., $\widehat{\by}=\by$). A larger deviation from the reference line (i.e., Bias(Slope)) indicates a greater systematic bias.
In addition, to determine if there is a linear relationship between the residuals and the predicted values, we calculated the correlation between them (i.e., $\cor(\widehat{\by},\hat{\epsilon})$). Moreover, to measure the overall prediction accuracy, we also compare the Root Mean Square Error (RMSE). To further investigate the presence of systematic bias around the tail areas, we calculated the mean prediction error (MPE), i.e., $\frac{1}{n}\sum_{i=1}^n(\widehat{\by}_i-\by_i)$ for observations falling below the lower quartile (i.e., $<Q_1$) and above the upper quartile (i.e., $>Q_3$). This measure evaluates whether the model systematically over-predicts (as indicated by a positive value) or under-predicts (as indicated by a negative value) the actual response values.

\begin{table*}[ht!]
\centering
\begin{tabular}{lllrrrrrrrrr}
  \hline
& & & KRR-UP & LASSO-UP  & KRR & LASSO & XGBoost & RF & NN & SVR\\ 
  \hline
\multirow{10}{*}{Linear}&\multirow{5}{*}{Train} &Bias (Slope) & \bf{0.0021} & 0.0038 & 0.0771 & 0.1761 & 0.0138 & 0.2795 & 0.0111 & 0.1555 \\ 
&&$\cor(\widehat{\by},\hat{\epsilon})$ & \bf{-0.0100} & -0.0113 & -0.3214 & -0.5712 & -0.6604 & -0.9295 & -0.1591 & -0.6615 \\ 
&&RMSE &  0.4272 & 0.5407 & 0.3897 & 0.5029 & \bf{0.0345} & 0.4940 & 0.1139 & 0.3846 \\ 
&&MPE ($<Q_1$)& \bf{0.0070} & 0.0098 & 0.1613 & 0.3646 & 0.0283 & 0.5749 & 0.0218 & 0.2875 \\ 
&&MPE ($>Q_3$)&-0.0066 & \bf{-0.0054} & -0.1612 & -0.3647 & -0.0242 & -0.5717 & -0.0208 & -0.2837 \\ 
   \cmidrule{2-11}
&\multirow{5}{*}{Test}&Bias (Slope)& 0.0350 & \bf{0.0138} & 0.1147 & 0.1939 & 0.5708 & 0.6589 & 0.1781 & 0.4033 \\ 
&&$\cor(\widehat{\by},\hat{\epsilon})$ & -0.1075 & \bf{-0.0416} & -0.3453 & -0.5717 & -0.7543 & -0.9366 & -0.3467 & -0.7519 \\ 
&&RMSE&0.5571 & 0.5906 & \bf{0.5509} & 0.5606 & 1.2564 & 1.1702 & 0.8531 & 0.8897 \\
&&MPE ($<Q_1$)& 0.0819 & \bf{0.0371} & 0.2461 & 0.4159 & 1.1790 & 1.3712 & 0.3237 & 0.7993 \\ 
&&MPE ($>Q_3$) & -0.0685 & \bf{-0.0289} & -0.2283 & -0.3973 & -1.1923 & -1.3640 & -0.3735 & -0.7898 \\ 
\hline
\multirow{10}{*}{Nonlinear 1}&\multirow{5}{*}{Train} &Bias (Slope) & \bf{0.0106} & 0.0285 & 0.2354 & 0.2709 & 0.0088 & 0.1869 & 0.0222 & 0.1370 \\ 
&&$\cor(\widehat{\by},\hat{\epsilon})$ & \bf{-0.0296} & -0.0616 & -0.6051 & -0.6555 & -0.4766 & -0.8782 & -0.1771 & -0.6448 \\ 
&&RMSE & 0.9933 & 1.2174 & 0.9872 & 1.0476 & \bf{0.0499} & 0.5403 & 0.3120 & 0.5395 \\ 
&&MPE ($<Q_1$)& -0.0570 & -0.0110 & 0.6789 & 0.7865 & \bf{0.0019} & 0.5530 & 0.0523 & 0.3380 \\ 
&&MPE ($>Q_3$)& -0.0693 & -0.1436 & -0.7967 & -0.9211 & \bf{-0.0479} & -0.6274 & -0.0769 & -0.4700 \\ 
   \cmidrule{2-11}
&\multirow{5}{*}{Test}&Bias (Slope)& \bf{0.0330} & 0.0455 & 0.2490 & 0.2748 & 0.2004 & 0.4304 & 0.1134 & 0.3411 \\ 
&&$\cor(\widehat{\by},\hat{\epsilon})$ & \bf{-0.0787} & -0.0945 & -0.5845 & -0.6302 & -0.4935 & -0.8894 & -0.2299 & -0.7047 \\ 
&&RMSE & 1.1415 & 1.3577 & 1.0895 & 1.1142 & \bf{1.0395} & 1.2458 & 1.2941 & 1.2431 \\ 
&&MPE ($<Q_1$)& \bf{-0.0079} & 0.0348 & 0.7269 & 0.8135 & 0.5127 & 1.3428 & 0.1783 & 1.0021 \\ 
&&MPE ($>Q_3$) &   -0.0693 & -0.1436 & -0.7967 & -0.9211 & \bf{-0.0479} & -0.6274 & -0.0769 & -0.4700 \\ 
\hline
\multirow{10}{*}{Nonlinear 2}&\multirow{5}{*}{Train} &Bias (Slope) & \bf{0.0034} & 0.0127 & 0.3248 & 0.4111 & 0.0113 & 0.2437 & 0.0167 & 0.1066 \\ 
&&$\cor(\widehat{\by},\hat{\epsilon})$ & \bf{-0.0099} & -0.0248 & -0.6504 & -0.7684 & -0.4031 & -0.9105 & -0.1473 & -0.5546 \\ 
&&RMSE &  2.7550 & 3.6517 & 2.4102 & 2.5808 & \bf{0.1422} & 1.2964 & 0.5343 & 0.9294 \\ 
&&MPE ($<Q_1$)& -0.0712 & 0.0055 & 1.9579 & 2.5048 & \bf{0.0091} & 1.4999 & 0.0904 & 0.5941 \\ 
&&MPE ($>Q_3$)& -0.1630 & -0.1896 & -2.1060 & -2.6291 & \bf{-0.1209} & -1.5085 & -0.1245 & -0.6557 \\ 
  \cmidrule{2-11}
&\multirow{5}{*}{Test}&Bias (Slope)& \bf{0.0274} & 0.0470 & 0.3369 & 0.4228 & 0.2680 & 0.5092 & 0.0552 & 0.2029 \\ 
&&$\cor(\widehat{\by},\hat{\epsilon})$ &  \bf{-0.0506} & -0.0754 & -0.6360 & -0.7523 & -0.5583 & -0.9167 & -0.1945 & -0.5926 \\ 
&&RMSE & 2.9646 & 3.9132 & 2.5810 & 2.7364 & 2.3465 & 2.7143 & \bf{1.4159} & 1.6647 \\ 
&&MPE ($<Q_1$)& \bf{0.1021} & 0.3057 & 2.0708 & 2.6290 & 1.4863 & 3.1748 & 0.3455 & 1.2029 \\ 
&&MPE ($>Q_3$) & -0.1733 & \bf{-0.1672} & -2.0830 & -2.5938 & -1.7674 & -3.1212 & -0.3047 & -1.1574 \\ 
  \hline
\end{tabular}
\caption{ Results of the synthetic experiments (low noise level) described in \Cref{Sec:Simulation}. The performance of methods is evaluated based on bias in slope, correlation between the predicted response and the residual, RMSE, and MPEs. The best-ranked method for each measure is highlighted in bold.}
\label{tab:sim_low}
\end{table*}

\begin{table*}[ht!]
\centering
\begin{tabular}{lllrrrrrrrrr}
  \hline
& & & KRR-UP & LASSO-UP  & KRR & LASSO & XGBoost & RF & NN & SVR\\ 
  \hline
\multirow{10}{*}{Linear}&\multirow{5}{*}{Train} &Bias (Slope) & \bf{0.0137} & 0.0209 & 0.2141 & 0.3996 & 0.0172 & 0.3083 & 0.0127 & 0.2673 \\ 
&&$\cor(\widehat{\by},\hat{\epsilon})$ & \bf{-0.0312} & -0.0383 & -0.5097 & -0.7388 & -0.6528 & -0.9229 & -0.2330 & -0.7174 \\ 
&&RMSE &  0.9330 & 1.1544 & 0.7670 & 0.9906 & \bf{0.0488} & 0.6152 & 0.1004 & 0.6828 \\ 
&&MPE ($<Q_1$)& \bf{0.0402} & 0.0622 & 0.5029 & 0.9302 & 0.0405 & 0.7081 & 0.0306 & 0.5766 \\ 
&&MPE ($>Q_3$)& -0.0338 & -0.0458 & -0.5020 & -0.9295 & -0.0350 & -0.7102 & \bf{-0.0308} & -0.5750 \\ 
   \cmidrule{2-11}
&\multirow{5}{*}{Test}&Bias (Slope)& 0.1106 & \bf{0.0580} & 0.3047 & 0.4416 & 0.6323 & 0.7170 & 0.3488 & 0.5372 \\ 
&&$\cor(\widehat{\by},\hat{\epsilon})$ & -0.1774 & \bf{-0.0914} & -0.5170 & -0.7356 & -0.7511 & -0.9254 & -0.4622 & -0.7796 \\ 
&&RMSE& 1.1800 & 1.2631 & 1.0871 & \bf{1.1056} & 1.5556 & 1.4331 & 1.3938 & 1.2729 \\ 
&&MPE ($<Q_1$)& 0.2782 & \bf{0.1535} & 0.7276 & 1.0568 & 1.4985 & 1.6925 & 0.8406 & 1.2401 \\ 
&&MPE ($>Q_3$) & -0.2424 & \bf{-0.1308} & -0.6917 & -1.0124 & -1.4326 & -1.6470 & -0.7523 & -1.2111 \\ 
\hline
\multirow{10}{*}{Nonlinear 1}&\multirow{5}{*}{Train} &Bias (Slope) &  0.0114 & \bf{0.0003} & 0.1436 & 0.2099 & 0.0059 & 0.1291 & 0.0986 & 0.0846 \\
&&$\cor(\widehat{\by},\hat{\epsilon})$ & -0.0194 & \bf{-0.0025} & -0.3122 & -0.4256 & -0.2931 & -0.5189 & -0.2219 & -0.3105 \\ 
&&RMSE & 0.7859 & 0.7896 & 0.6140 & 0.7918 & \bf{0.0345} & 0.3982 & 0.5929 & 0.3632 \\ 
&&MPE ($<Q_1$)& -0.0533 & -0.1232 & 0.4423 & 0.6673 & \bf{0.0042} & 0.4115 & 0.2949 & 0.2248 \\ 
&&MPE ($>Q_3$)& -0.0778 & -0.0603 & -0.5108 & -0.7428 & \bf{-0.0341} & -0.4503 & -0.3474 & -0.3143 \\ 
   \cmidrule{2-11}
&\multirow{5}{*}{Test}&Bias (Slope)& 0.0511 & \bf{0.0124} & 0.1569 & 0.2136 & 0.1920 & 0.3037 & 0.1172 & 0.2050 \\ 
&&$\cor(\widehat{\by},\hat{\epsilon})$ & -0.0875 & \bf{-0.0245} & -0.3122 & -0.4215 & -0.3357 & -0.5287 & -0.2351 & -0.3697 \\ 
&&RMSE &  0.8224 & 0.9131 & 0.6966 & 0.8422 & 0.9466 & 0.9572 & \bf{0.6862} & 0.7748 \\
&&MPE ($<Q_1$)& \bf{0.0792} & -0.1014 & 0.4981 & 0.7157 & 0.5621 & 1.0552 & 0.3293 & 0.6396 \\
&&MPE ($>Q_3$) &  -0.1570 & \bf{-0.0393} & -0.5483 & -0.7157 & -0.6799 & -1.0408 & -0.4214 & -0.7503 \\ 
\hline
\multirow{10}{*}{Nonlinear 2}&\multirow{5}{*}{Train} &Bias (Slope) & \bf{0.0047} & 0.0171 & 0.3448 & 0.4352 & 0.0126 & 0.2474 & 0.0256 & 0.1340 \\
&&$\cor(\widehat{\by},\hat{\epsilon})$ & \bf{-0.0128} & -0.0309 & -0.6608 & -0.7801 & -0.4092 & -0.8981 & -0.1712 & -0.5573 \\  
&&RMSE &  2.9982 & 3.9416 & 2.5683 & 2.7428 & \bf{0.1582} & 1.3600 & 0.7170 & 1.1799 \\ 
&&MPE ($<Q_1$)&-0.0302 & \bf{0.0180} & 2.1418 & 2.7261 & 0.0157 & 1.5539 & 0.1280 & 0.7932 \\ 
&&MPE ($>Q_3$)& \bf{-0.0509} & -0.0801 & -2.1788 & -2.7379 & -0.1308 & -1.5214 & -0.1935 & -0.7857 \\ 
  \cmidrule{2-11}
&\multirow{5}{*}{Test}&Bias (Slope)& \bf{0.0253} & 0.0408 & 0.3541 & 0.4446 & 0.2897 & 0.5185 & 0.0929 & 0.2392 \\ 
&&$\cor(\widehat{\by},\hat{\epsilon})$ &  \bf{-0.0451} & -0.0661 & -0.6493 & -0.7674 & -0.5582 & -0.9078 & -0.2237 & -0.5958 \\ 
&&RMSE & 3.1570 & 4.2054 & 2.7238 & 2.8894 & 2.5889 & 2.8532 & 2.0635 & \bf{2.0003} \\ 
&&MPE ($<Q_1$)& \bf{-0.0192} & 0.1990 & 2.1853 & 2.8023 & 1.6676 & 3.3383 & 0.5245 & 1.4316 \\ 
&&MPE ($>Q_3$) & -0.2412 & \bf{-0.2225} & -2.2599 & -2.8025 & -1.9560 & -3.2311 & -0.6341 & -1.4871 \\ 
  \hline
\end{tabular}
\caption{Results of the synthetic experiments (high noise level) described in \Cref{Sec:Simulation}. The performance of methods is evaluated based on bias in slope, correlation between the predicted response and the residual, RMSE, and MPEs. The best-ranked method for each measure is highlighted in bold.}
\label{tab:sim_high}
\end{table*}

The results averaged over 100 replications are provided in \Cref{tab:sim_low,tab:sim_high} for low and high levels of noise respectively. As shown in these tables, LASSO-UP and KRR-UP outperform the competing methods, exhibiting only minimal systematic bias in slope across all settings and in both the training and testing datasets, with bias values being consistently less than 0.1, except for KRR-UP in the testing set under the linear setting.

In contrast, all benchmark methods exhibit systematic biases in both training and testing datasets with a biased slope, high linear correlation between predicted values and residuals, and systematic bias in the tail areas. Specifically, these models tend to produce a positive bias for responses under the lower quantiles, indicating that the predicted values are generally larger than the observed values. Conversely, responses above the upper quantiles typically result in negative biases, suggesting an underestimation of the observed values. In addition, both XGBoost and random forest are subject to overfitting, as evidenced by their superior performance on the training set, particularly in terms of RMSE, compared to their performance on the testing set. The issue of overfitting with these ML methods is more pronounced in the linear case, as evidenced by a bias in slope greater than 0.55. This implies that the predicted values tend to be closer to the overall mean, indicating that the SBMR occurs. 

Due to the `bias-variance' tradeoff, the performance criteria involving the variance measures (e.g., RMSE) of LASSO-UP and KRR-UP may not be optimal as the constraints can be considered as regularization terms (ref Hastie book). This is because our objective function does not globally minimize the squared error loss by incorporating the constraints designed to correct the systematic bias. Consequently, this may result in an increase in variance, which can be considered as the cost to eliminate the systematic bias.

\section{Real-life example (Estimating Brain Age)} \label{Sec:Real}
The systematic bias of predicted outcomes by machine learning regression has been commonly reported in the field of neuroimaging research, particularly in the context of estimating brain age to quantify an individual's brain condition by regressing chronological age on neuroimaging measurements. In such cases, older individuals tend to be estimated with a younger brain age, while younger individuals tend to be estimated with an older brain age.

We applied the proposed methods to brain age computation for two neuroimaging studies: The Lifespan Human Connectome Project Aging Study (\texttt{HCP-A}, \url{https://www.humanconnectome.org/study/hcp-lifespan-aging}), and UK Biobank (\texttt{UKBB}, \url{https://www.ukbiobank.ac.uk/}) data. 
In both the \texttt{HCP-A} and \texttt{UKBB} datasets, we utilized fractional anisotropy (FA), a measure of white matter integrity, measured across various brain regions as the neuroimaging predictors. The dimensions of the predictors are $p=64$ for \texttt{HCP-A} and $p=39$ for \texttt{UKBB}, respectively. We used chronological age as the response variable. The age range for the \texttt{HCP-A} dataset is 36-90 years, and for the \texttt{UKBB} dataset, it is 45-82 years. The number of samples for which neuroimaging measurements are available is 662 for \texttt{HCP-A} and 36,856 for \texttt{UKBB}.

We trained the model on randomly sampled subsets from the entire dataset—300 samples for \texttt{HCP-A} and 500 for \texttt{UKBB}. We then evaluated model performance using the unseen testing set, which consists of the remaining data for \texttt{HCP-A} and another 500 randomly selected samples from the remaining data for \texttt{UKBB}. We compared the performance of our method with other ML methods. The results from 100 replications are summarized in \Cref{tab:Real_result}.

\begin{table*}[ht!]
\centering
\begin{tabular}{lllrrrrrrrr}
  \hline
 & & & KRR-UP & LASSO-UP & KRR & LASSO  & XGBoost & RF & NN & SVR\\ 
  \hline
\multirow{10}{*}{\texttt{HCP-A}} & \multirow{5}{*}{Training} &Bias (Slope) & 0.0360 & 0.0546 & 0.4884 & 0.3991 & 0.0257 & 0.2025 & \bf{0.0049} & 0.2513 \\ 
& &$\cor(\widehat{\by},\hat{\epsilon})$ & \bf{-0.0421} & -0.0892 & -0.7668 & -0.6741 & -0.4800 & -0.7486 & -0.0687 & -0.5919 \\
& &RMSE &12.4646 & 8.7541 & 9.0563 & 8.3976 & 0.7662 & 3.8452 & \bf{0.8723} & 6.0374 \\ 
& &MPE ($<Q_1$)& 2.9523 & 2.8410 & 10.1944 & 8.5987 & 0.4522 & 4.1654 & \bf{0.1127} & 5.3767 \\ 
& &MPE ($>Q_3$)& 0.3543 & \bf{-0.0258} & -8.3816 & -6.6933 & -0.5207 & -3.4804 & -0.0785 & -4.1737 \\
\cmidrule{2-11}
&  \multirow{5}{*}{Testing} &Bias (Slope) &  \bf{0.0836} & 0.1011 & 0.5014 & 0.4218 & 0.4557 & 0.5094 & 0.3666 & 0.4422 \\ 
& &$\cor(\widehat{\by},\hat{\epsilon})$ & \bf{-0.0946} & -0.1466 & -0.7602 & -0.6652 & -0.6138 & -0.7523 & -0.3751 & -0.6677 \\ 
& &RMSE &13.0586 & 9.9780 & 9.4402 & \bf{9.0679} & 10.6410 & 9.6918 & 14.1034 & 9.5163 \\ 
& &MPE ($<Q_1$)& 3.9517 & \bf{3.8288} & 10.4991 & 9.1117 & 9.2023 & 10.4802 & 6.9275 & 8.9889 \\ 
& &MPE ($>Q_3$)& \bf{-0.3372} & -0.7189 & -8.5892 & -7.0786 & -8.2942 & -8.7487 & -7.6181 & -7.7641 \\ 
   \hline
\multirow{10}{*}{\texttt{UKBB}} & \multirow{5}{*}{Training} &Bias (Slope) & \bf{0.0112} & 0.0548 & 0.6575 & 0.6964 & 0.1125 & 0.2729 & 0.0365 & 0.3935 \\ 
& &$\cor(\widehat{\by},\hat{\epsilon})$ & \bf{-0.0099} & -0.0473 & -0.8406 & -0.8679 & -0.6376 & -0.8381 & -0.1892 & -0.7118 \\
& &RMSE & 10.4443 & 8.9972 & 5.9655 & 6.1140 & \bf{1.3448} & 2.4839 & 1.4287 & 4.2211 \\ 
& &MPE ($<Q_1$)& \bf{0.9762} & 1.2967 & 7.1503 & 7.5326 & 1.0664 & 2.9075 & 0.3301 & 4.5794 \\ 
& &MPE ($>Q_3$)& 0.7251 & \bf{0.0154} & -6.2527 & -6.6633 & -1.2362 & -2.6515 & -0.4204 & -3.4132 \\ 
\cmidrule{2-11}
&  \multirow{5}{*}{Testing} &Bias (Slope) &  \bf{0.0570} & 0.1423 & 0.6713 & 0.7046 & 0.6368 & 0.6728 & 0.5707 & 0.6146 \\ 
& &$\cor(\widehat{\by},\hat{\epsilon})$ & \bf{-0.0456} & -0.1129 & -0.8374 & -0.8647 & -0.7299 & -0.8386 & -0.4369 & -0.7612 \\ 
& &RMSE &10.7556 & 9.8626 & \bf{6.1274} & 6.2263 & 6.6727 & 6.1303 & 10.0439 & 6.1825 \\ 
& &MPE ($<Q_1$)& \bf{1.3715} & 2.0788 & 7.2935 & 7.6138 & 6.5043 & 7.1610 & 5.1442 & 6.7661 \\ 
& &MPE ($>Q_3$)& \bf{0.3159} & -0.6783 & -6.3269 & -6.6889 & -6.4051 & -6.5001 & -6.4815 & -5.7160 \\ 
   \hline
\end{tabular}
\caption{Results of the real data application. The best-ranked method for each measure is highlighted in bold. }
\label{tab:Real_result}
\end{table*}

\begin{figure*}
    \centering
    \includegraphics[width=0.99\linewidth]{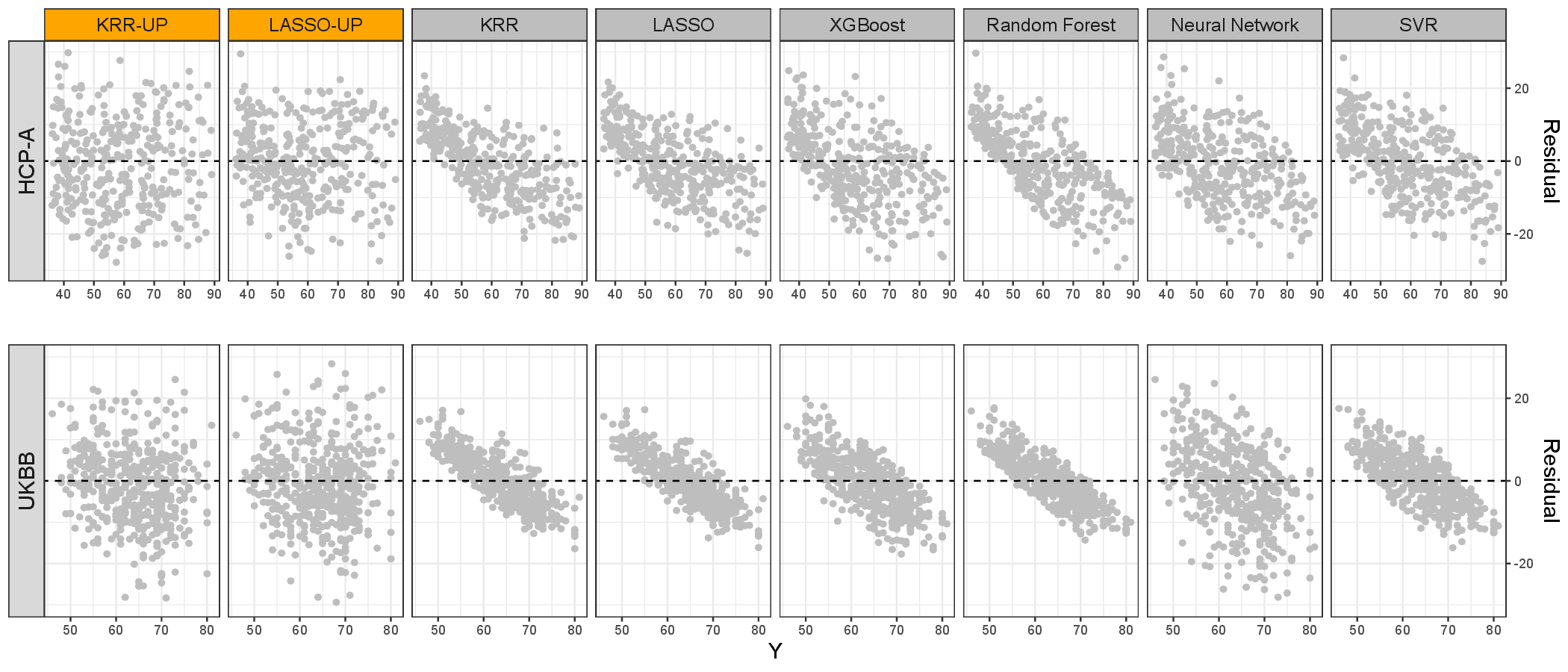}
    \caption{This Scatter plot, representing one result from 100 replications, shows the observed response $\by_i$ ($x$-axis) and the residuals $\widehat{\by}_i-\by_i$ ($y$-axis) from various methods on the testing set. The reference dashed line represents $\by=0$. This figure demonstrates that the residuals from the proposed methods (KRR-UP and LASSO-UP) in the orange panels are randomly distributed, regardless of the value of the response variable. In contrast, other machine learning regression models in the gray panels exhibit a pattern in their errors, showing positive errors for low response values and negative errors for high response values.}
    \label{fig:Real_residual}
\end{figure*}

Overall, KRR-UP outperforms other methods, with the only exceptions being lower tail bias in the \texttt{HCP-A} data, where the proposed LASSO-UP performs the best with only a small margin, and RMSE for both datasets, where conventional methods perform well. The small bias in slope (all less than 0.1) of KRR-UP indicates that the predicted values are closely aligned with the observed values. Additionally, the correlation between residuals and response values is significantly weaker for the proposed methods compared to other competing methods. This is also illustrated in \Cref{fig:Real_residual}, which displays a scatter plot of one sample result from 100 replications. This lack of a linear trend indicates that the proposed method corrects systematic biases, leading to more reliable predictions across the range of data. This removal of systematic bias is further evidenced by the small biases observed in the tail areas, below the first quartile ($<Q_1$) and above the third quartile ($>Q_3$).

As noted, the proposed method's reduction in bias may lead to increased variance in model predictions, potentially raising the RMSE. Consequently, our method is not optimal in terms of RMSE, illustrating a trade-off between bias and variance. However, to correct the systematic bias and avoid the additional bias by post-bias correction, the increased RMSE of KRR-UP may be acceptable.

\section{Conclusion}\label{Sec:Conclusion}

The bias-variance trade-off is a fundamental concept in machine learning research. Since the objective function in most machine learning regression models aims to minimize Mean Squared Error (MSE) without explicitly constraining prediction bias, the predictive function may tolerate some bias in exchange for minimizing the overall objective function. In this paper, we show that this bias is often systematic across various machine learning prediction models, which can negatively impact the primary utility of machine learning regression analysis: decision-making and forecasting.


To correct the systematic bias in machine learning regression models, we propose a solution by introducing two-sided constraints to the commonly used objective function. By implementing these constraints, we developed two machine learning regression models with unbiased predictions: LASSO-UP and KRR-UP, both of which can handle high-throughput predictors (i.e., $p>n$). Furthermore, we demonstrate through theoretical analysis and data experiments that LASSO-UP and KRR-UP effectively correct bias and produce unbiased predictions compared to various commonly used machine learning regression models. Lastly, our real-world application of brain age prediction using neuroimaging data showcases unbiased predictions, addressing a long-standing gap in brain age research.



We conclude this paper by acknowledging its limitations and suggesting future research directions. While our correction methods effectively mitigate systematic bias, the proposed models still face the bias-variance trade-off, leading to a higher mean squared error compared to benchmark machine learning regression models. Future research will extend these corrections to other popular machine learning regression models, such as Random Forest, XGBoost, and others, with the goal of achieving optimal performance across all corrected methods. The preferred model will provide unbiased predictions with minimal RMSE. 


The \texttt{R} package implementing LASSO-UP and KRR-UP, as used in this paper, is publicly available online at \url{https://github.com/hwiyoungstat/SBMR}.





\appendix
\subsection{Proof of Proposition 1}

Let $\ddot{\by}_i$ denote an unbiased prediction $\by_i$. Recall that $R^2 = 1-\dfrac{\sum_{i=1}^n(\by_i-\ddot{\by}_i)^2}{\sum_{i=1}^n(\by_i-\overline{\by})^2}$, and $\sum_{i=1}^n (\by_i-\ddot{\by}_i)^2 = \sum_{i=1}^n(\by_i-\overline{\by})^2 \times (1-R^2)$. Without loss of generality, we have $\mathbb{E}\left[\sum_{i=1}^n (\by_i-\ddot{\by}_i)^2\right] = (n-1)\sigma^2 (1-R^2)$.\\
\noindent Suppose $\widetilde{\by}_i = \ddot{\by}_i-c\ddot{\by}_i$ is the linear trend of the systematic bias, then following holds.
\begin{align*}
    \mathbb{E}\left[\sum_{i=1}^n (\by_i-\widetilde{\by}_i)^2\right] &= \mathbb{E}\left[\sum_{i=1}^n\left(\by_i-\mathbb{E}(\widetilde{\by}_i) + \mathbb{E}(\widetilde{\by}_i)-\widetilde{\by}_i\right)^2\right]\\
    &= \mathbb{E}\left[\sum_{i=1}^n (\by_i-\mathbb{E}(\widetilde{\by}_i))^2 + \sum_{i=1}^n(\mathbb{E}(\widetilde{\by}_i)-\widetilde{\by}_i)^2\right.\\ & \hspace{.8cm}\left.+ \sum_{i=1}^n2(\by_i-\mathbb{E}(\widetilde{\by}_i))(\mathbb{E}(\widetilde{\by}_i)-\widetilde{\by}_i)\right]\\
    &= \mathbb{E}\sum_{i=1}^n (c\by_i)^2 + \mathbb{E}\sum_{i=1}^n [(1-c)(\by_i-\ddot{\by}_i)]^2 \\ & \hspace{.8cm} + 2\mathbb{E}\sum_{i=1}^n(\by_i-\mathbb{E}(\widetilde{\by}_i))(\mathbb{E}(\widetilde{\by}_i)-\widetilde{\by}_i)\\
    &= (n-1)c^2\sigma^2 + (n-1)(1-c)^2\sigma^2(1-R^2).
\end{align*}

It suffices to verify the condition where $\mathbb{E}\left[\sum_{i=1}^n (\by_i-\ddot{\by}_i)^2\right] > \mathbb{E}\left[\sum_{i=1}^n (\by_i-\widetilde{\by}_i)^2\right]$: 
\begin{align*}
    (1-R^2) &> c^2 + (1-c)^2(1-R^2)\\
    (1-R^2) &> 1-2c+2c^2-R^2+2cR^2-c^2R^2\\
    2c-2c^2-2cR^2+c^2R^2 &> 0\\
    1-c-R^2+cR^2/2 &> 0 \rightarrow \left(1-\frac{R^2}{2}\right)c < 1-R^2
\end{align*}
Thus, we have $\mathbb{E}\left[\sum_{i=1}^n (\by_i-\ddot{\by}_i)^2\right] > \mathbb{E}\left[\sum_{i=1}^n (\by_i-\widetilde{\by}_i)^2\right]$, when $c < \frac{1-R^2}{1-R^2/2}$. $\square$

Therefore, machine learning regression models with the objective function to minimize mean squared error naturally incorporate the linear bias $\widetilde{y}_i = \widehat{y}_i-c\widehat{y}_i$  in their prediction. 

\subsection{Proof of Theorem 1}
The proof is straightforward. The two constraints imply $\sum_{i\in\mathbb{I}_{<}} (\by_i-\widetilde{\by}_i)=\sum_{i\in\mathbb{I}_{<}}(\by_i-\ddot{\by}_i+c\ddot{\by}_i)=0$, and $\sum_{i\in\mathbb{I}_{>}} (\by_i-\widetilde{\by}_i)=\sum_{i\in\mathbb{I}_{>}}(\by_i-\ddot{\by}_i+c\ddot{\by}_i)=0$. 
\begin{align}
    0&=\sum_{i\in\mathbb{I}_{<}} (\by_i-\widetilde{\by}_i)+\sum_{i\in\mathbb{I}_{>}} (\by_i-\widetilde{\by}_i)\notag\\ 
    &=\sum_{i\in\mathbb{I}_{<}}(\by_i-\ddot{\by}_i+c\ddot{\by}_i)+\sum_{i\in\mathbb{I}_{>}}(\by_i-\ddot{\by}_i+c\ddot{\by}_i)\notag\\ 
    &= \sum_{i=1}^n(\by_i-\ddot{\by}_i) + \sum_{i=1}^n c\ddot{\by}_i \label{eq:pf_thm1}
\end{align}
Since $\ddot{\by}$ is the unbiased prediction of $\by$, taking expectation of the first term in \eqref{eq:pf_thm1} is 0 (i.e., $\mathbb{E} \sum_{i=1}^n(\by_i-\ddot{\by}_i)=0$.  To ensure $c \mathbb{E}\sum_{i=1}^n\ddot{\by}_i =0$ hold for all cases of $\sum_{i=1}^n \widehat{\by}_i$, the constant $c$ must be 0.

\bibliographystyle{IEEEtran}
\bibliography{reference.bib}

\begin{thebibliography}{10}
\providecommand{\url}[1]{#1}
\csname url@samestyle\endcsname
\providecommand{\newblock}{\relax}
\providecommand{\bibinfo}[2]{#2}
\providecommand{\BIBentrySTDinterwordspacing}{\spaceskip=0pt\relax}
\providecommand{\BIBentryALTinterwordstretchfactor}{4}
\providecommand{\BIBentryALTinterwordspacing}{\spaceskip=\fontdimen2\font plus
\BIBentryALTinterwordstretchfactor\fontdimen3\font minus \fontdimen4\font\relax}
\providecommand{\BIBforeignlanguage}[2]{{%
\expandafter\ifx\csname l@#1\endcsname\relax
\typeout{** WARNING: IEEEtran.bst: No hyphenation pattern has been}%
\typeout{** loaded for the language `#1'. Using the pattern for}%
\typeout{** the default language instead.}%
\else
\language=\csname l@#1\endcsname
\fi
#2}}
\providecommand{\BIBdecl}{\relax}
\BIBdecl

\bibitem{ESL}
T.~Hastie, R.~Tibshirani, and J.~Friedman, \emph{The Elements of Statistical Learning}, 2nd~ed.\hskip 1em plus 0.5em minus 0.4em\relax Springer New York, NY, 2009.

\bibitem{Zhang:2012_RFbias}
\BIBentryALTinterwordspacing
G.~Zhang and Y.~Lu, ``Bias-corrected random forests in regression,'' \emph{Journal of Applied Statistics}, vol.~39, no.~1, pp. 151--160, 2012. [Online]. Available: \url{https://doi.org/10.1080/02664763.2011.578621}
\BIBentrySTDinterwordspacing

\bibitem{Belitz:2021}
\BIBentryALTinterwordspacing
K.~Belitz and P.~Stackelberg, ``Evaluation of six methods for correcting bias in estimates from ensemble tree machine learning regression models,'' \emph{Environmental Modelling \& Software}, vol. 139, p. 105006, 2021. [Online]. Available: \url{https://www.sciencedirect.com/science/article/pii/S1364815221000499}
\BIBentrySTDinterwordspacing

\bibitem{SMITH:2019}
\BIBentryALTinterwordspacing
S.~M. Smith, D.~Vidaurre, F.~Alfaro-Almagro, T.~E. Nichols, and K.~L. Miller, ``Estimation of brain age delta from brain imaging,'' \emph{NeuroImage}, vol. 200, pp. 528--539, 2019. [Online]. Available: \url{https://www.sciencedirect.com/science/article/pii/S1053811919305026}
\BIBentrySTDinterwordspacing

\bibitem{Butler:2021}
\BIBentryALTinterwordspacing
E.~R. Butler, A.~Chen, R.~Ramadan, T.~T. Le, K.~Ruparel, T.~M. Moore, T.~D. Satterthwaite, F.~Zhang, H.~Shou, R.~C. Gur, T.~E. Nichols, and R.~T. Shinohara, ``Pitfalls in brain age analyses,'' \emph{Human Brain Mapping}, vol.~42, no.~13, pp. 4092--4101, 2021. [Online]. Available: \url{https://onlinelibrary.wiley.com/doi/abs/10.1002/hbm.25533}
\BIBentrySTDinterwordspacing

\bibitem{Treder:2021}
\BIBentryALTinterwordspacing
M.~S. Treder, J.~P. Shock, D.~J. Stein, S.~du~Plessis, S.~Seedat, and K.~A. Tsvetanov, ``Correlation constraints for regression models: Controlling bias in brain age prediction,'' \emph{Frontiers in Psychiatry}, vol.~12, 2021. [Online]. Available: \url{https://www.frontiersin.org/journals/psychiatry/articles/10.3389/fpsyt.2021.615754}
\BIBentrySTDinterwordspacing

\bibitem{Wang:2023}
H.~Wang, M.~S. Treder, D.~Marshall, D.~K. Jones, and Y.~Li, ``A skewed loss function for correcting predictive bias in brain age prediction,'' \emph{IEEE Transactions on Medical Imaging}, vol.~42, no.~6, pp. 1577--1589, 2023.

\bibitem{Tibshirani:1996}
\BIBentryALTinterwordspacing
R.~Tibshirani, ``Regression shrinkage and selection via the lasso,'' \emph{Journal of the Royal Statistical Society. Series B (Methodological)}, vol.~58, no.~1, pp. 267--288, 1996. [Online]. Available: \url{http://www.jstor.org/stable/2346178}
\BIBentrySTDinterwordspacing

\bibitem{Gaines:2018}
\BIBentryALTinterwordspacing
B.~R. Gaines, J.~Kim, and H.~Zhou, ``Algorithms for fitting the constrained lasso,'' \emph{Journal of Computational and Graphical Statistics}, vol.~27, no.~4, pp. 861--871, 2018. [Online]. Available: \url{https://doi.org/10.1080/10618600.2018.1473777}
\BIBentrySTDinterwordspacing

\bibitem{Nocedal:2006}
J.~Nocedal and S.~J. Wright, \emph{Numerical Optimization}, 2nd~ed.\hskip 1em plus 0.5em minus 0.4em\relax New York: Springer, 2006.

\bibitem{ScSm:2002}
B.~Sch{\"o}lkopf and A.~J. Smola, \emph{Learning with Kernels:Support vector machines, Regularization, Optimization, and Beyond}.\hskip 1em plus 0.5em minus 0.4em\relax Cambridge, MA, USA: MIT Press, 2002.

\bibitem{Wahba:1990}
\BIBentryALTinterwordspacing
G.~Wahba, \emph{Spline Models for Observational Data}.\hskip 1em plus 0.5em minus 0.4em\relax Society for Industrial and Applied Mathematics, 1990. [Online]. Available: \url{https://epubs.siam.org/doi/abs/10.1137/1.9781611970128}
\BIBentrySTDinterwordspacing

\bibitem{Hofmann:2008}
\BIBentryALTinterwordspacing
T.~Hofmann, B.~Sch{\"o}lkopf, and A.~J. Smola, ``{Kernel methods in machine learning},'' \emph{The Annals of Statistics}, vol.~36, no.~3, pp. 1171 -- 1220, 2008. [Online]. Available: \url{https://doi.org/10.1214/009053607000000677}
\BIBentrySTDinterwordspacing

\bibitem{Chen:2016}
\BIBentryALTinterwordspacing
T.~Chen and C.~Guestrin, ``Xgboost: A scalable tree boosting system,'' ser. KDD '16.\hskip 1em plus 0.5em minus 0.4em\relax New York, NY, USA: Association for Computing Machinery, 2016, p. 785–794. [Online]. Available: \url{https://doi.org/10.1145/2939672.2939785}
\BIBentrySTDinterwordspacing

\bibitem{Breiman2001}
\BIBentryALTinterwordspacing
L.~Breiman, ``Random forests,'' \emph{Machine Learning}, vol.~45, no.~1, pp. 5--32, 2001. [Online]. Available: \url{https://doi.org/10.1023/A:1010933404324}
\BIBentrySTDinterwordspacing

\bibitem{Schmidhuber:2015}
\BIBentryALTinterwordspacing
J.~Schmidhuber, ``Deep learning in neural networks: An overview,'' \emph{Neural Networks}, vol.~61, pp. 85--117, 2015. [Online]. Available: \url{https://www.sciencedirect.com/science/article/pii/S0893608014002135}
\BIBentrySTDinterwordspacing

\bibitem{SVR:1996}
\BIBentryALTinterwordspacing
V.~Vapnik, S.~Golowich, and A.~Smola, ``Support vector method for function approximation, regression estimation and signal processing,'' in \emph{Advances in Neural Information Processing Systems}, M.~Mozer, M.~Jordan, and T.~Petsche, Eds., vol.~9.\hskip 1em plus 0.5em minus 0.4em\relax MIT Press, 1996, pp. 281--287. [Online]. Available: \url{https://proceedings.neurips.cc/paper_files/paper/1996/file/4f284803bd0966cc24fa8683a34afc6e-Paper.pdf}
\BIBentrySTDinterwordspacing

\bibitem{Smola:2004}
\BIBentryALTinterwordspacing
A.~J. Smola and B.~Sch{\"{o}}lkopf, ``A tutorial on support vector regression,'' \emph{Statistics and Computing}, vol.~14, no.~3, pp. 199--222, Aug. 2004. [Online]. Available: \url{https://doi.org/10.1023/B:STCO.0000035301.49549.88}
\BIBentrySTDinterwordspacing

\bibitem{Friedman:1991}
\BIBentryALTinterwordspacing
J.~H. Friedman, ``{Multivariate Adaptive Regression Splines},'' \emph{The Annals of Statistics}, vol.~19, no.~1, pp. 1 -- 67, 1991. [Online]. Available: \url{https://doi.org/10.1214/aos/1176347963}
\BIBentrySTDinterwordspacing

\end{thebibliography}







\end{document}